\documentclass{article}
\usepackage{ijcai20}

\usepackage{times}

\usepackage{soul}
\usepackage{url}
\usepackage[hidelinks]{hyperref}
\usepackage[utf8]{inputenc}
\usepackage[small]{caption}
\usepackage{graphicx}
\usepackage{amsmath}
\usepackage{amsthm}
\usepackage{amssymb}
\usepackage{booktabs}
\usepackage{multirow}
\usepackage{algorithm}
\usepackage{algorithmic}
\DeclareUnicodeCharacter{2212}{-}
\DeclareUnicodeCharacter{FB01}{fi}

\renewcommand{\thefootnote}{\fnsymbol{footnote}}

\title{Cross-denoising Network against Corrupted Labels in \\ Medical Image Segmentation with Domain Shift}

\author{
    Qinming Zhang$^{2*}$, Luyan Liu$^{1*\dagger}$, Kai Ma$^1$, Cheng Zhuo$^2$, Yefeng Zheng$^{1\dagger}$
    \affiliations
    $^1$Tencent Jarvis Lab, Shenzhen, China \\
    $^2$Zhejiang University, Hangzhou, China 
    \emails
      \{qinmingzhang, czhuo\}@zju.edu.cn,
      \{luyanliu, kylekma, yefengzheng\} @tencent.com
}

\begin{document}
\maketitle
\let\thefootnote\relax\footnotetext{* Authors contributed equally.}
\let\thefootnote\relax\footnotetext{$\dagger$ Corresponding authors.}

\begin{abstract}
Deep convolutional neural networks (DCNNs) have contributed many breakthroughs in segmentation tasks, especially in the field of medical imaging. However, \textit{domain shift} and \textit{corrupted annotations}, which are two common problems in medical imaging, dramatically degrade the performance of DCNNs in practice. In this paper, we propose a novel robust cross-denoising framework using two peer networks to address domain shift and corrupted label problems with a peer-review strategy. Specifically, each network performs as a mentor, mutually supervised to learn from reliable samples selected by the peer network to combat with corrupted labels. In addition, a noise-tolerant loss is proposed to encourage the network to capture the key location and filter the discrepancy under various noise-contaminant labels.
To further reduce the accumulated error, we introduce a class-imbalanced cross learning using most confident predictions at the class-level. Experimental results on REFUGE and Drishti-GS datasets for optic disc (OD) and optic cup (OC) segmentation demonstrate the superior performance of our proposed approach to the state-of-the-art methods.
\end{abstract}

\section{Introduction}

The performance of current deep convolutional neural networks (DCNNs) highly depends on two assumptions: (1) training and test data are drawn from the same feature space with the same distribution; and (2) training data is associated with accurate annotations. However, the performance of established DCNN models usually degrades when tested on unseen data, especially when there exists significant appearance difference between training (source domain) and test (target domain) data, which is referred as the domain shift problem. To mitigate such problem, tremendous domain adaptation (DA) methods have been proposed \cite{tzeng2017adversarial,chen2018domain,tsai2018learning} Nevertheless, most of the current DA solutions assume that the ground-truth labels in training data are flawless, thus ignore an inevitable problem—labels may be corrupted in the real world \cite{han2018coteaching}. This unique challenge inspires us to consider one problem: ``How can we learn a robust domain adaptive model from data with noisy annotations?”.

\begin{figure}[t] 
	\centering
	\includegraphics[width=1\linewidth]{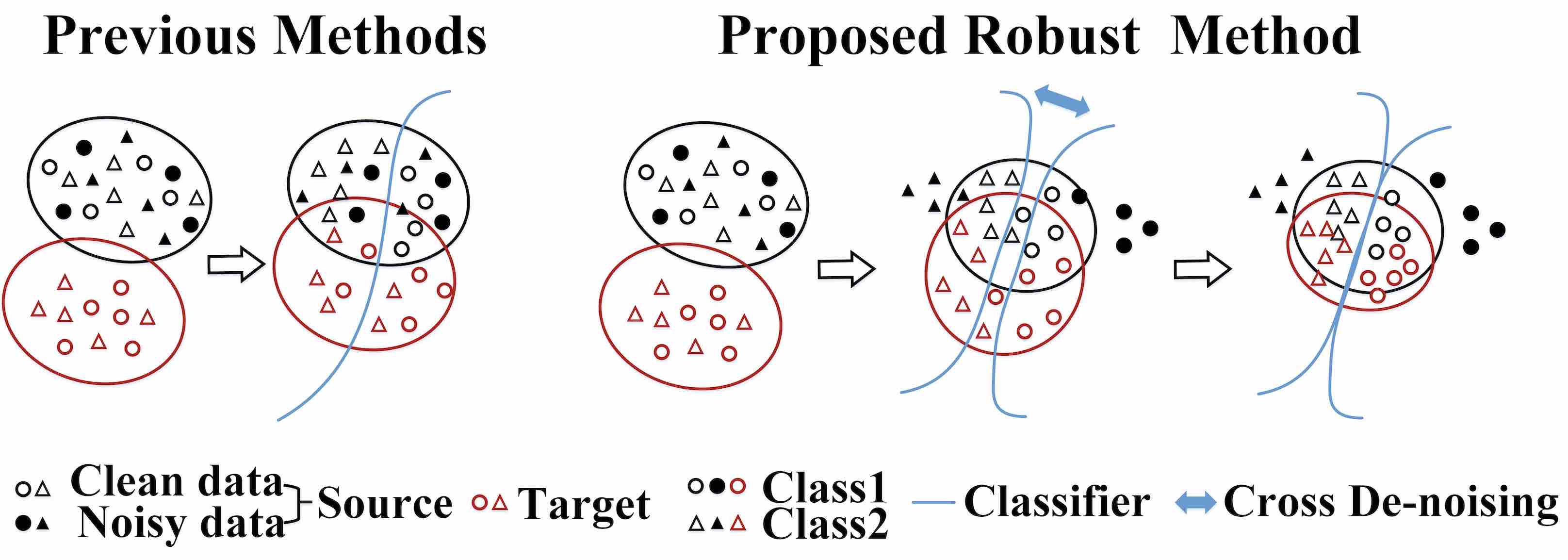} 
	\caption{Schematic diagrams of the noise-label effects during domain adaptation. Compared to previous solutions, our proposed cross-denoising network can better align the source  (black ellipse) and target (red ellipse) domains by filtering out the noise-contaminant labels.}\label{fig: introduction} 
\end{figure}

Domain shift is a common problem in the field of medical imaging, since images are obtained from special medical devices, where different imaging modalities or even different settings of the same device could introduce significant variations in images. Recently, many approaches are emerging to address the domain shift problem in image segmentation. Li \textit{et al.} \shortcite{li2019bidirectional} proposed a bidirectional learning method with self-supervised learning to learn a better segmentation model and in return improve the image translation model. Vu \textit{et al.} \shortcite{vu2018advent} proposed an entropy-based adversarial training approach targeting structure adaptation from source domain to target domain. Additionally, manual annotation with a pixel-level accuracy is indeed inefficient and error-prone. The wrong-labelled samples, behaving as ``noise'', can potentially degrade the performance of DCNN, thus it is challenging to learn from data with domain shift and noisy annotations.

Aiming to alleviate the above problems, we propose a robust cross-denoising framework that is resilient to noisy annotations and domain shift. We design two different networks playing roles as peer reviewers to selectively learn from the data with reliable clean labels and adaptively correct the training error. Furthermore, we introduce a class-imbalanced self-learning
strategy to estimate the most reliable labels for the target domain. Fig. \ref{fig: introduction} illustrates the main idea of previous DA methods and our proposed robust cross-denoising method. We evaluate the cross-denoising model against the state-of-the-art methods on the REFUGE dataset \cite{tz6e-r977-19} and the Drishti-GS dataset \cite{Sivaswamy2015ACR}. In this nutshell, our main contributions of this paper are summarized as follows: \\ 1) We firstly (to the best of our knowledge) propose a robust learning method against noisy labels in medical image segmentation with domain shift.\\2) We propose a cross-reviewing framework that identifies high-quality data and a noise-tolerant loss to focus on the noise-free part in noisy labels, which can significantly reduce the negative effects of noisy labels and boost the performance of two peer networks.\\ 3) We introduce a class-imbalanced cross learning
strategy in an iterative cross-training procedure. The presented novel approach enables generating target labels with higher confidence and accuracy. \\ 4) We demonstrate that this robust framework achieves state-of-the-art on optic disc (OD) and optic cup (OC) segmentation tasks with domain shift and noisy labels.

\begin{figure}[t] 
	\centering
	\includegraphics[width=0.98\linewidth]{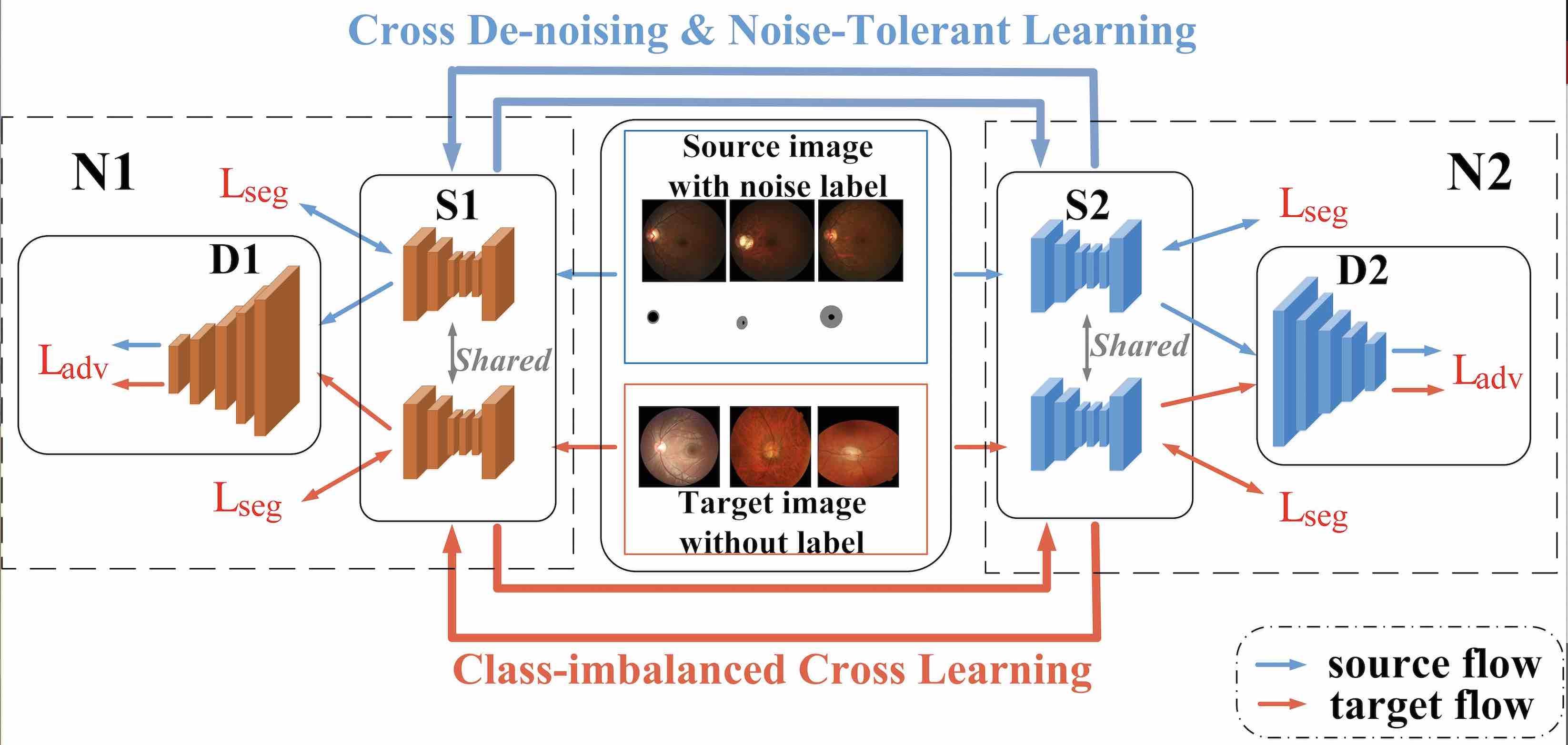} 
	\caption{Pipeline of our proposed unsupervised domain adaptation framework.}\label{fig: framework} 
\end{figure}

\section{Related Works}
\paragraph{Domain Adaptation.} Recently, there are increasing studies proposed to address the domain shift problem with domain adaptation techniques. Many approaches have achieved promising performance on natural image datasets. For instance, Chang \textit{et al.} \shortcite{chang2019structure} proposed a DICE framework, disentangling the representation of an image into a domain-invariant structure component and a domain-specific texture component, to advance domain adaptation for semantic segmentation. Aiming to address the problem of semantic inconsistency incurred by global feature alignment, Luo \textit{et al.} \shortcite{luo2018taking} took a close look at the category-level joint distribution and aligned each class with an adaptive adversarial loss. For medical image segmentation, Dou \textit{et al.} \shortcite{dou2018unsupervised} proposed a plug-and-play domain adaptation module (DAM) by adapting the source and target domains in the feature space, to solve the cardiac structure segmentation problem across different modalities.  The latest study on medical data that is closely related to our work is \cite{Wang_2019}, which presented a patch-based output space adversarial learning framework to jointly segment the OD and OC from different fundus image datasets. However, all these existing DA methods rely on training data with clean annotations whose performance would be degraded dramatically once the annotations are corrupted or ambiguous.

\paragraph{Noisy Labeling.} Training DCNNs with the presence of corrupted labels is a challenging task, which has attracted numerous researchers working towards solutions. Among those works, one of the representative methods is \cite{jiang2017mentornet}, which proposed a MentorNet to supervise the training of a StudentNet and select samples that were probably correct. Another work, \cite{han2018coteaching}, introduced a co-teaching strategy to robustly train the deep neural networks under noisy supervision. For medical imaging, Xue \textit{et al.} \shortcite{xue2019robust} proposed an iterative learning strategy for imperfectly labeled skin lesion image classfication, combating the lacking of clean annotated medical data. Existing approaches on robust learning about noisy labeling are mostly focused on the image classification task, which leaves segmentation with corrupted labels an unsolved problem. In this paper, we provide a novel solution to address the medical image segmentation task with both domain shift and contaminated label problems at the same time.

\begin{figure}[t] 
	\centering
	\vspace{0pt}
	\includegraphics[width=1\linewidth]{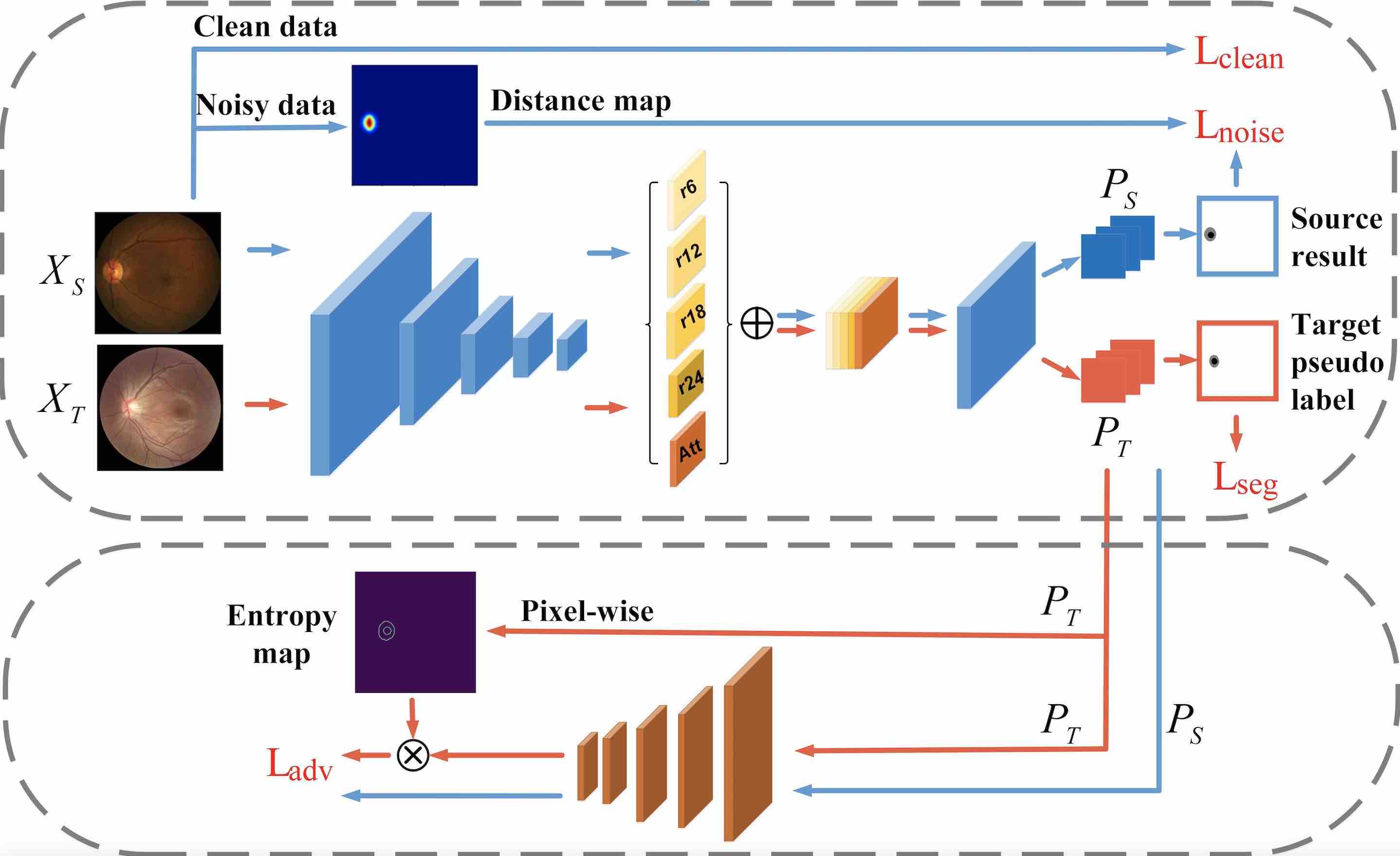} 
	\caption{Architecture of the sub-network N1.}\label{fig: segmentation network} 
\end{figure}

\section{Method}\label{Method}
The overall architecture of our proposed robust cross-denoising network is shown in Fig. \ref{fig: framework}, which consists of two different networks working as peer reviewers in an unsupervised domain adaption fashion. In this section, we firstly illustrate the architecture of the proposed cross-denoising network. Then,  a robust cross-denoising learning algorithm is designed to learn an accurate and robust model from contaminated labels. Last but not least, we propose a noise-tolerant loss and a class-imbalanced cross learning strategy to learn critical information from corrupted labels, which are elaborated in Sections 3.2 and 3.3, respectively.

\subsection{Robust Cross-Denoising Network}
\paragraph{Network Architecture.} As shown in Fig. \ref{fig: framework},  our proposed cross-denoising network (CD-Net) consists of two different networks (i.e., N1, N2), both of which include a segmentation network (resp. S1, S2) and a discriminator (resp. D1, D2). N1 and N2, playing roles as two experts, can generate different decision boundaries, thus there should be differences in their learning abilities and opinions. For network N1, we follow the spirit of  DeepLabv2 \cite{chen2016deeplab} architecture with ResNet101 \cite{he2015deep} as backbone to achieve initial segmentation results.  For network N2, in order to learn discriminative features different from N1, we adopt DeepLabv3+ architecture with MobileNetv2 as backbone \cite{chen2018encoderdecoder}. To boost the segmentation ability of N1 to the same level of N2, we design a novel spatial pyramid pooling (ASSP) structure \cite{chen2018encoderdecoder} with multi-attention mechanism \cite{fu2018dual} for N1, which is shown in Fig. \ref{fig: segmentation network}, so as to enhance the feature expression ability and enrich the multi-scale information of the network. As regard to D1 and D2, we adopt the same architecture, which is a 5-layer fully convolutional network. D1 and D2 are trained to distinguish between the source prediction and the target prediction by adversarial learning, and guide the segmentation network to focus on the local structure similarity. In the testing stage, we only use N1 to generate the final segmentation results.

\paragraph{Robust Cross-denoising Learning.}
Our proposed robust cross-denoising algorithm is shown in Algorithm 1. With a subset data $C_{k}$ (step 3), we train two different networks $N_{1}$ and $N_{2}$ to select a propotion of samples with small training loss (steps 6 and 7). Based on the observation of deep networks \cite{yu2019does}, easy cases can be learned firstly, and then the networks gradually fit to the hard cases with the number of epochs increased. Therefore, in a noisy dataset, the network learns clean and easy parts of data in the early stage, and thus has the abilities to filter out noisy pattern using loss values. The number of filtered samples is controlled by remember rate $\gamma$, which increases (step 13) utill it filters out all the potential noisy data. Since the learning and filtering ability of networks is not strong enough in early epochs, the remember rate is initialized with a small value and becomes larger when epochs increase. 
After then, the selected high-quality data from one network is fed into its peer network as reliable knowledge to update parameters (steps 9 and 10). Since two networks have different structure and learning abilities, they can filter different types of error introduced by noisy labels. Although the error caused by noisy labels is propagated back from one network itself, the other network can adaptively correct the training error with a prediction disagreement between two networks. Based on such peer-review strategy, each network selects its small-loss samples as the high-quality data, and updates its peer network by such clean samples to further reduce the training error. 

\begin{algorithm}[t]
	\caption{Cross-denoising Algorithm}
	\textbf{Input}: The source domain training set $X_S$, model parameters  $\Omega^1$ and $\Omega^2$, noise ratio $\beta$, remember rate $\gamma$, learning rate $\delta$, epoch number $T$, iteration numbers $K$ and $M$.
	\begin{algorithmic}[1]
		\FOR{$t=1$ to $T$}
		\STATE Initialize $\gamma = \gamma_0 $
		\STATE Randomly sample a subset $C_k$ from $X_S$ \\
		// Sample $\gamma$ of small-loss instances from each \\ // peer network as high-quality data: \\
		\FOR{$k=1$ to $K$}
		\STATE Compute the overall loss function by Eq. (3)\\
		\STATE ${C_k^1} = argmin_{\hat C_k: |\hat C_k| \geq  \gamma |C_k|}  \mathcal{L}_{our} (\hat C_k | \omega = 0) $  and\\
		\STATE  ${C_k^2} = argmin_{\hat C_k: |\hat C_k| \geq \gamma |C_k|}  \mathcal{L}_{our} (\hat C_k | \omega = 0) $\\
		// Exchange the high-quality data $C_k^1$ and $C_k^2$ to  \\// update the peer network $N_2$ and $N_1$
		\FOR{$i=1$ to $M$}
		\STATE Update $\Omega^1 = \Omega^1- \delta \nabla \mathcal{L}_{our} (\hat C_k^2)$ \\
		\STATE Update $\Omega^2= \Omega^2- \delta \nabla \mathcal{L}_{our} (\hat C_k^1)$ \\ 
		\ENDFOR
		\ENDFOR 
		\STATE Update $\gamma =
		min\left\{ \frac {t \times {(1-\beta)}}{T},1-\beta \right\}; $
		\ENDFOR
		\\
		\STATE Output model parameters $\Omega^1$ and $\Omega^2$. 
	\end{algorithmic}
\end{algorithm}

\paragraph{Overall Training Objective.} The proposed cross-denoising domain adaptation network includes two loss functions: noise-tolerant segmentation loss and noise-robust adverserial loss.
Among the high-quality data selected by the network, not all of them are clean data, some of them may be mixed with noisy data. In order to learn from clean labels and corrupted labels, respectively, we seperate the data into two groups, i.e., data with reliable label (clean data) and noisy label (noisy data) based on the prediction confidence. For the noise-tolerant segmentation loss $\mathcal{L}_{seg}$, it consists of segmentation loss for clean data and corrupted data which is shown in Eq. (1) and will be elaborated further in Section 3.2. When the instance is grouped in clean data, the noise-filtering segmentation loss is equal to ${L}_{clean}$ ($\omega=0$); otherwise, it is formulated as ${L}_{noise}$ ($\omega=1$)

\begin{equation}
\begin{aligned}
\mathcal{L}_{seg}(p,y)&= (1-\omega)\mathcal{L}_{clean}+\omega\mathcal{L}_{noise}.
\end{aligned}
\end{equation} 

Since the unlabeled data in the target domain can be regarded as the extreme case of data with noisy labels. The direct prediction in the target domain is usually inaccurate and noisy, which affects the convergence and generalization of adversarial learning. To maximize prediction certainty, an ``entropy map" is multiplied by the predictions for the target domain image, which increases the loss weight for the pixels with inaccurate and noisy estimated labels, and reduces the loss weight for accurate and clean estimated labels.  The entropy map of the predicted result in the target domain is defined as:
$\mathcal F(X_i)=-\sum_{i=1}^{h\times w\times c}p_i\log(p_i)$. 
We adopt the entropy map as an indicator to weight the noise-robust adversarial loss ${L}_{adv}$, which is defined as :
\begin{equation}
\begin{aligned}
\mathcal{L}_{adv}(X_S,X_T)=-{E}[log(D(G(X_S)))]\\
-{E}[(\lambda_{entr}\mathcal F(X_T)+\epsilon)+log(1-D(G(X_T)))],
\end{aligned}
\end{equation} 
where $\lambda_{entr}$ is the weight parameter corresponding to information entropy map, and $\epsilon$ is to ensure the stability of the training process in the case of a small $\mathcal F(X_T)$. 

The training objective function for our proposed noise-robust segmentation method can be formulated as the following min-max criterion:

\begin{equation}
\begin{aligned}
\mathcal{L}_{our}=\min_{\mathbf{G}}\max_\mathbf{D}\mathcal{L}_{seg}(p,y)+\lambda_{adv}\mathcal{L}_{adv},
\end{aligned}
\end{equation}
where $\lambda_{adv}$ denotes the hyperparameter controlling the weights of the adversarial loss, which is empirically set as 0.001. 

\subsection{Learning from Corrupted Labels} 
Since clean data can obtain small loss while  remaining corrupted data large loss, we use a hybrid segmentation loss composed of the common cross-entropy loss and the Dice coefficient loss, which are shown in the second term and third term in Eq. (4), respectively. Given images $X$ in source domain $S$ and target domain $T$ with the size $h$ (height) by $w$ (width), set $c$ as the number of classes, the clean data segmentation loss can be concretized as:

 \begin{equation}
 \begin{aligned}
 \mathcal{L}_{clean}(p,y)&= 1 -\lambda_1 \sum_{i=1}^{h\times w \times c}{y_i}\log(p_i) \\
 &- \lambda_2 \dfrac {2 \sum_{i=1}^{h\times w \times c}y_i\cdot  p_i} {y_i^2 + p_i^2},
 \label{equation: 1} 
 \end{aligned}
 \end{equation} 
 where $p_i = \mathbf{G}(X) \in \mathbb{R}^{(h\times w\times c)}$ is the softmax output of the segmentation network, and $y_i$ is the ground-truth. $\lambda_1$ and $\lambda_2$ are the weights to improve network training, which are empirically set as 0.05 and 1,  respectively. For the corrupted data, since the incorrect annotations are mostly around the boundary in practice, the annotations inside the segmented regions are more reliable. Inspired by this observation, we propose to selectively learn from noisy labels. The noisy data segmentation loss shown in Eq. (5) prevents the network from overfitting the noisy pixels while keeping the ability to learn from the reliable pixels in noisy data:

 \begin{equation}
 \begin{aligned}
 \mathcal{L}_{noise}(p,y)&= 1 -\lambda_1 \sum_{i=1}^{h\times w \times c}\mathcal {B}(y_i)\log(p_i) \\
 &- \lambda_2 \dfrac {2 \sum_{i=1}^{h\times w \times c}\mathcal{B}(y_i)\cdot  p_i} {\mathcal{B}(y_i)^2 + p_i^2},\vspace{-5pt}
 \end{aligned}
 \end{equation} 
 
 \begin{equation}
 \begin{aligned}
 \mathcal {B}(y_i^c)=
 exp\big(-\dfrac {(max{(D(y_i^c))}-D(y_i^c))^2} {2 \times \delta^2}\big), \ c=1,2,...C, \\
 \end{aligned}
 \end{equation} 
 where ${\mathcal B}(y_i)$ denotes the boundary distance map.
 As boundary is generally more vulnerable to noise in a medical image, we calculate the distance $D(y_i)$ to the nearest boundary for each pixel $y_i$, and get the maximum of $D(y_i)$ in class-level region, namely $max{(D(y_i))}$. $\delta$ represents the standard deviation, which can be defined as $\delta = \left.max{(D(y_i))} \middle /2.58\right.$ because 99\% of Gaussian distribution is in range ${(−2.58 \times \delta, 2.58 \times \delta)}$.  With respect to $\mathcal B(y_i^c)$, the center of the region in each class has a larger value, and the closer to the boundary, the smaller the value. Such noise-tolerant loss $\mathcal {L}_{noise}$ encourages the network to capture the key location in the center and filter out the discrepancy in the boundary under various noise-contaminant labels.
\vspace{-5pt} 
\subsection{Class-imbalanced Cross Learning}
In case of unsupervised domain adaptation with ambiguous labels, it is more challenging to esimate the results accurately. Using the predictions of the learned model as the latent variables for the target image, which is called ``Pseudo Label" (PL),  is an alternative way to solve such intractable problem. Because of the presence of corrupted labels and different class distributions, the predictions are not robust to the noisy disturbance and the levels of prediction difficulty among classes are different. The vanilla self-learning does not take such issue into consideration, and selects pseudo labels using universal confidence for each class. We propose a class-imbalanced cross learning strategy to solve this issue (shown in Algorithm 2), in which we select the pseudo labels with most confident predictions at class-level and feed them into the peer network to be resistant to noise. Specifically, we use the trained segmenter to predict the latent target labels (step 2),  and rank the prediction values of each category (step 4) to select the pixels with value greater than the confidence threshold as pseudo labels (step 7). The generated pseudo labels are fed into the peer discriminator (step 9) to adaptively correct the adversarial error of companion, which is robust to the noise based on above discussions in Section 3.2. The algorithm is conducted in an iterative cross training procedure.

\begin{algorithm}[t]
	\caption{Class-Imbalanced Cross Learning Algorithm}
	\textbf{Input}: The target domain training set $X_T$, threshold $q^c$, iteration number $I$ and class number $C$.
	\begin{algorithmic}[1]
		\FOR{$i=1$ to $I$}
		\STATE  Calculate the prediction $p(i)$ in the target domain
		\FOR{$c=1$ to $C$}
		\STATE $p^c(i) = sort(p^c(i), order = descending)$ and \\
		\STATE $thresh(c) = p^c(i)[q^c \times length(p^c(i))]$ \\ 
		\ENDFOR
		\STATE $PL = argmax(p)_{[p^c>thresh(c)]}$
		\ENDFOR
		\STATE Exchange the pseudo-labels $PL^1$ and $PL^2$ to the peer networks $N_2$ and $N_1$.
	\end{algorithmic}
\end{algorithm}

\begin{table*}[h]
	\centering
	\setlength{\tabcolsep}{1mm}{
		\begin{tabular}{|c|c|c|c|c|c|c|c|c|c|c|}
			\hline
			\multirow{2}{*}{\begin{tabular}[c]{@{}c@{}}\small {Noise}\\   \small {Level}\end{tabular}} & \multirow{2}{*}{\small {Pretrain}} & \multirow{2}{*}{\begin{tabular}[c]{@{}c@{}}\small {Noise}\\   \small {Ratio}\end{tabular}} & \multicolumn{2}{c|}{\small{BDL} \tiny{\cite{li2019bidirectional}}} & \multicolumn{2}{c|}{\small{pOSAL}  \tiny{\cite{Wang_2019}}} & \multicolumn{2}{c|}{\small{BEAL} \tiny{\cite{wang2019boundary}}} & \multicolumn{2}{c|}{\small {Proposed}} \\ \cline{4-11} 
			&  &  & \small{$REF$} & \small{$DGS$} & \small{$REF$} & \small{$DGS$} & \small{$REF$} & \small{$DGS$} & \small{$REF$} & \small{$DGS$} \\ \hline
			\multirow{7}{*}{\small {Low}} & \multirow{4}{*}{\small {With}} & \small 0 & \small (94.6, 87.4) & \small (94.6, 82.8) & \small (94.9, 88.7) & \small (95.5, 84.5) & \small (93.3, 83.1)  & \small (93.8, 85.0) & \small (\textbf{95.3}, \textbf{89.4}) & \small (\textbf{96.1}, \textbf{85.9})\\  
			&  & \small 0.1 & \small (94.8, 88.7) & \small (94.6, 83.1) & \small (\textbf{95.4}, 88.0) & \small (\textbf{95.3}, 83.6) & \small (93.1, 82.0) & \small (93.4, 83.1) & \small (95.1, \textbf{89.3}) & \small (95.1, \textbf{83.8}) \\ 
			&  & \small 0.5 & \small (94.9, 89.0) & \small (94.3, 80.8) & \small (94.9, 85.9) & \small (94.9, 80.9) & \small (90.2, 80.5) & \small (93.1, 82.1) & \small (\textbf{95.4}, \textbf{89.6}) & \small (\textbf{95.8}, \textbf{84.2}) \\ 
			&  & \small 0.9 & \small (94.2, 86.8) & \small (94.0, 82.6) & \small (94.5, 85.8) & \small (94.8, 80.6) & \small (87.7, 80.5) & \small \small (93.2, 78.0)  & \small (\textbf{95.3}, \textbf{89.4}) & \small (\textbf{95.1}, \textbf{82.9}) \\ \cline{2-11} 
			
			& \multirow{3}{*}{\small W/O} & \small 0.1 & \small (94.2, 86.7) & \small (92.6, 82.5) & \small (94.1, 87.9) & \small (94.8, 81.4) & \small (92.7, 77.1) & \small (92.4, 78.2) & \small  (\textbf{94.7}, \textbf{88.4}) & \small (\textbf{95.1}, \textbf{83.8})\\ 
			&  & \small 0.5 & \small (93.2, 86.0) & \small (87.6, 81.1) & \small (94.0, 85.0) & \small  (92.6, 78.4) & \small (87.8, 75.8)  & \small (87.3, 78.0) & \small (\textbf{94.8}, \textbf{86.7}) & \small  (\textbf{94.4}, \textbf{83.7}) \\ 
			&  & \small 0.9 & \small (90.6, 76.5) & \small (85.6, 80.3) & \small (92.5, 83.6) & \small (88.6, 77.7) & \small (82.8, 69.1) & \small (83.5, 74.7) & \small (\textbf{94.1}, \textbf{84.6}) & \small (\textbf{92.7}, \textbf{83.4}) \\ \hline
			
			\multirow{6}{*}{\small High} & \multirow{3}{*}{\small With} 
			& \small 0.1 & \small (94.7, 83.3) & \small (94.1, 81.1) & \small (94.7, 86.5) & \small (92.6, 81.3) & \small (92.4, 81.8) & \small (92.6, 80.1) & \small (\textbf{95.1}, \textbf{89.0}) & \small (\textbf{94.6}, \textbf{83.0})\\ 
			&  & \small 0.5 & \small (91.6, 79.6) & \small (87.9, 68.5) & \small (85.8, 75.6) & \small (91.3, 78.7) & \small (89.1, 77.2) & \small (92.4, 74.9) & \small (\textbf{93.9}, \textbf{85.6}) & \small (\textbf{93.4}, \textbf{81.3}) \\ 
			&  & \small 0.9 & \small (90.2, 74.3) & \small (85.9, 65.1) & \small (84.5, 76.0) & \small (91.6, 76.3) & \small (75.9, 66.9) & \small (90.4, 73.5) & \small (\textbf{93.0}, \textbf{83.6}) & \small (\textbf{92.4}, \textbf{82.7})  \\ \cline{2-11} 
			
			& \multirow{3}{*}{\small W/O} & \small 0.1 & \small (89.5, 75.9) & \small (89.2, 72.3) & \small (88.2, 78.7) & \small (87.5, 59.0) & \small (91.2, 73.8) & \small (68.2, 56.5) & \small (\textbf{94.5}, \textbf{88.8}) & \small (\textbf{92.7}, \textbf{83.0}) \\ 
			&  &\small 0.5 & \small (85.8, 75.6) & \small (85.6, 66.9) & \small (83.9, 74.5) & \small (85.0, 54.4) & \small (86.9, 69.1) & \small (73.4, 59.6) & \small (\textbf{93.8}, \textbf{84.7}) & \small (\textbf{93.0}, \textbf{81.8}) \\ 
			&  & \small 0.9 & \small (84.7, 66.0) & \small (81.6, 68.7) & \small (79.0, 72.0) & \small (81.6, 56.3) & \small (77.8, 53.2) & \small (70.5, 50.7) & \small (\textbf{92.9}, \textbf{81.1}) & \small (\textbf{91.8}, \textbf{80.4}) \\ \hline
	\end{tabular}}
	\caption{Test accuracy of different methods from the REFUGE training set to the REFUGE validation set (REF) and the Drishti-GS test set (DGS), respectively. (*, *) represents (\small $DI_{disc}$, \small $DI_{cup}$).}
	\label{tab: Refuge}
\end{table*}

\section{Experiments}
\subsection{Datasets}
In this study, we verify our approach on two public optic disc (OD) and optic cup (OC) segmentation datasets, including the REFUGE challenge dataset \cite{tz6e-r977-19} and the Drishti-GS dataset \cite{Sivaswamy2015ACR} (Table \ref{tab: dataset}). We refer the  REFUGE training set as the source domain, the REFUGE validation set and Drishti-GS dataset  as the target domains 1 and 2, respectively. The source domain contains some ground truth labels and imperfect labels, while the target domain contains no labels. Each target domain is further split into a training set for unsupervised DA (ignoring the labels) and a test set. The source and target domain images are acquired by different scanners resulting in different color and texture characteristics of the images. Extensive experiments on these two public databases with different noise levels and noise ratios are conducted to verify the effectiveness of our proposed approach.

\begin{table}[h]
	\setlength{\tabcolsep}{0.6mm}{
		\begin{tabular}{ccccc}
			\toprule 
			\small{Domain} &  \small{Dataset}&  \small{Training}&  \small{Test} &  \small{Size}\\
			\midrule  
			\small{Source}& \small{REFUGE training set} & \small{400} & \small{None}& \small{$2124\times2056$}\\
			\small{Target1} & \small{REFUGE validation set} &  \small{300} & \small{100} & \small{$1634\times1634$}\\
			\small{Target2} & \small{Drishti-GS} &  \small{50} & \small{51} & \small{$2047\times1759$} \\
			\bottomrule 
	\end{tabular}}
	\caption{Summary of datasets used in the experiments.}
	\label{tab: dataset}
\end{table}
\vspace{-20pt}
\begin{figure}[t] 
	\centering
	\includegraphics[width=1\linewidth]{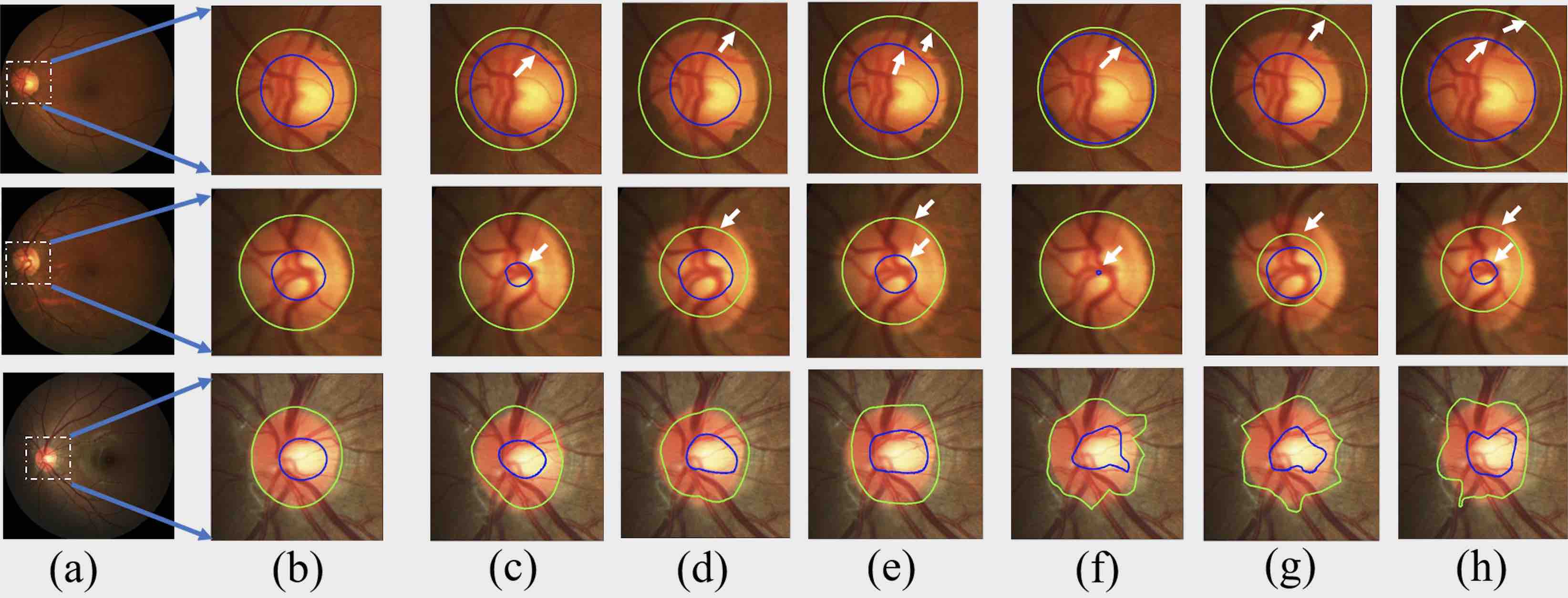} 
	\vspace{-20pt}
	\caption{Examples of generated noisy labels. Each row from (c) to (h) represents dilated, eroded
		and non-rigidly transformed labels respectively. (a) Original images, (b) ground-truth, (c)(d)(e) labels with low noise level, (f)(g)(h) labels with high noise level.}
	\label{fig: labels} 
	\vspace{-5pt}
\end{figure}

\begin{figure*}[t] 
	\centering
	\vspace{-10pt}
	\includegraphics[width=1\linewidth]{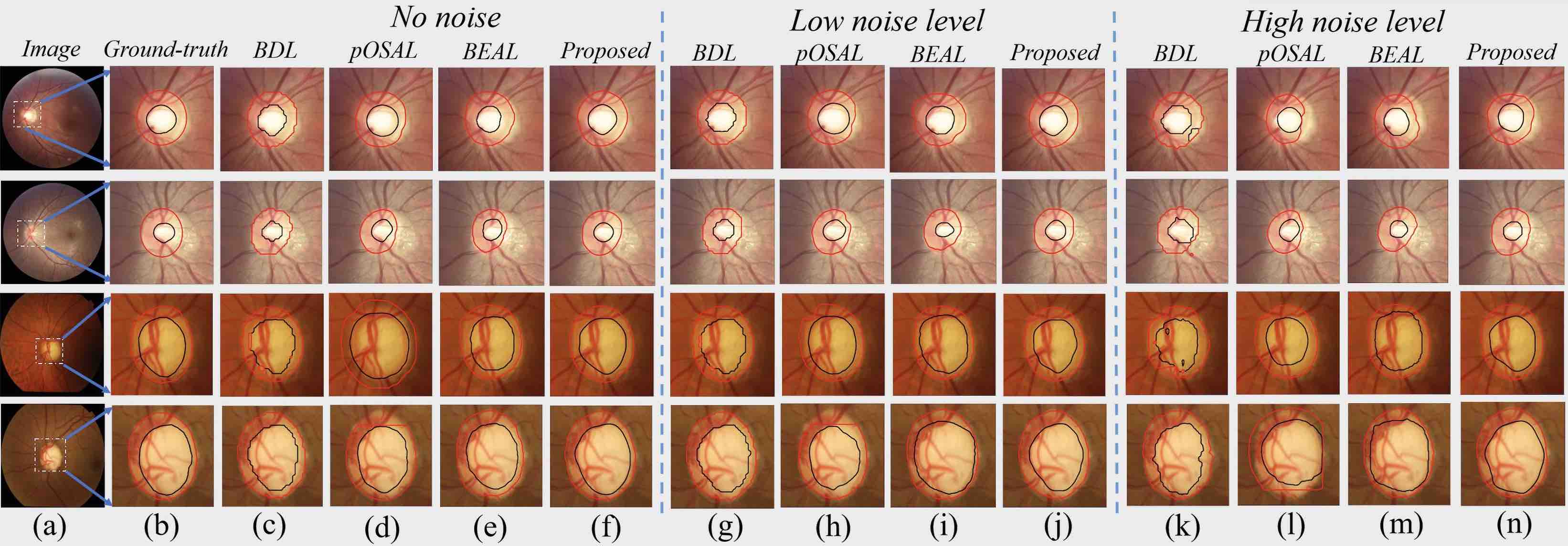} 
	\vspace{-20pt}
	\caption{Results of different methods for OD and OC segmentations with different noise levels at noise ratio of 0.5. Each row presents one typical example.  (a) Original image, (b)	 ground-truth, (c)(g)(k) BDL method \protect \cite{li2019bidirectional}, (d)(h)(l) pOAL meathod \protect \cite{Wang_2019}, (e)(i)(m) BEAL method \protect \cite{wang2019boundary}, and (f)(j)(n) our proposed method. Segmentation boundaries for OD and OC are shown in red and black, respectively.}\label{fig: results}
\end{figure*}

\vspace{22pt}
We generate three types of noisy labels as shown in Fig. \ref{fig: labels}: i) enlarge the label mask by dilation, ii) shrink the labels by erosion, and iii) deform the labels by elastic deformation. Varying the amount of dilation, erosion, and deformation, we generate corrupted noisy datasets with different noise levels, which are measured as function $\alpha = \sum_{i=1}^c {(1-DI_i)}$, where $DI_i$ is the Dice coefficient between generated noisy labels and ground-truth of class $i$. Specifically, we empirically set low noise level as $0.1 \le \alpha \le 0.4$ and high noise level as $0.4 \le \alpha \le 0.7$. We also set different noise ratio $\beta$ that represents portion of corrupted samples randomly selected from the training set, where $\beta \in {(0.1, 0.5, 0.9)}$.

\subsection{Implementation Details}
The proposed method is implemented using PyTorch on 4 Tesla P40 GPU with 96 GB memory in total. We use the Stochastic Gradient Descent optimizer with a momentum of 0.9 to train the segmentation network, and the Adam optimizer to train the discriminator.  The initial learning rates are $2.5\times10^{-4}$ and $1\times10^{-4}$ for the segmentation network and the discriminator, respectively.

\subsection{Quantitative Results}
We compare our proposed method with the state-of-the-art unsurpervised DA methods including BDL \cite{li2019bidirectional}, pOSAL \cite{Wang_2019}, and BEAL \cite{wang2019boundary} for the OD and OC segmentation on different noise levels and noise ratios. The Dice coefficients (DI) of OD and OC are used as evaluation criteria. 


\paragraph{Comparative Study.} Table \ref{tab: Refuge} presents the perfrormance comparison of all the methods transferring from the REFUGE training to the REFUGE validation and Drishti-GS test datasets with different noise levels and noise ratios. As for the REFUGE dataset (REF), we notice that the impact of label noise is not identical for all neural networks. On clean-annotated dataset, all methods work well and our proposed method achieve the best performance, with $DI_{disc}$ of 95.3 and  $DI_{cup}$ of 89.4. But as the noise ratio increases, the competitor methods have different degrees of degradation while our method can still maintain a stable and robust result. It is because we not only identify high-quality data effectively, but also avoid the error accumulation issue and assimilate the gains of clean data. Therefore, our method can reach higher performance and combat with harder cases. Furthermore, we observe that when using a pretrained model at low noise level, the performance shows no sign of declining at some cases. 
This indicates that the pretrained model can improve model robustness \cite{hendrycks2019using} and take the mild noise as a form of ``data augmentation", which relaxes the learning criterion and boosts the performance of competitors and our method. 
When training at high noise level, the performances of the competitor methods are declining sharply  with the increase of noise ratio. 
In contrast, our method can detect the most reliable data and learn from samples prone to be corrupted, thus we can learn more discriminative features and achieve better performance. More specifically,  in the hardest case of 0.9 noisy ratio, our method beats the best competitor pOSAL with 17.6\% $DI_{disc}$ and 12.6\% $DI_{cup}$ improvement when training from scratch. 

\begin{table}[h]
	\centering
	\setlength{\tabcolsep}{1mm}{
		\begin{tabular}{ccccccccc}
			\toprule
			\multicolumn{3}{c}{\small Strategy} & \multicolumn{2}{c}{\small 0.1}  & \multicolumn{2}{c}{\small 0.5}  & \multicolumn{2}{c}{\small 0.9} \\ 
			\small CD & \small CICL & \small NTL & \small $DI_{disc}$ & \small $DI_{cup}$ & \small $DI_{disc}$ & \small $DI_{cup}$ & \small $DI_{disc}$ & \small $DI_{cup}$  \\ \midrule
			&  &  & \small 94.6 & \small 87.7  & \small 83.8 & \small 72.6  & \small 80.1 & \small 71.5 \\ 
			\checkmark &  &  & \small 95.0 & \small 88.4  & \small 89.7 & \small 81.5  & \small 84.7 & \small 77.3 \\ 
			\checkmark & \checkmark&  & \small 95.1 & \small 88.9  & \small 92.6 & \small 83.6  & \small 88.9 & \small 80.3 \\ 
			\checkmark& \checkmark & \checkmark & \small \textbf{95.1} & \small \textbf{89.0} & \small \textbf{93.9} & \small \textbf{85.6} & \small \textbf{93.0} & \small \textbf{83.6} \\ 
			\bottomrule
	\end{tabular}}
	\caption{Test accuracy with different strategies from the REFUGE training set to the REFUGE validation set with noisy ratios at 0.1, 0.5, 0.9, respectively. CD refers to cross-denoising learning by two different networks. CICL is the class-imbalanced cross learning. NTL means the noise-tolerant loss.}
	\label{tab: ablation}
\end{table}

The results on the Drishti-GS dataset (DGS) have the similar trends as REF. Because the distributions of RFUGE and Drishti-GS datasets are quite different, the performance of competitors is in steep decline for the larger domain shift, while our method can alleviate such domain shift and learn from pseudo labels with high confidence. Concretely, at 0.9 noisy ratio, our method beats the best competitor
BDL with 12.5\% $DI_{disc}$ and 13.5\% $DI_{cup}$ improvement when training from scratch.
The qualitative testing results on the REFUGE and Drishti-GS datasets are visualized in Fig. \ref{fig: results}. In the case of no noise, the competitor methods can locate the approximate location but fail to generate accurate boundaries of OD and OC. In contrast, our method successfully localizes the OD and OC and generates more accurate boundaries. With noise added, the differences between the segmentation results of competitors and ground-truth become prominent, while our model can still achieve promising results and show its superiority over other methods. 

\paragraph{Ablation Study.} We also conduct a set of ablation experiments to investigate the effectiveness of each component as exhibited in Table \ref{tab: ablation}.
With the CD strategy, the performance has increased significantly, which validates that the module can gain from high-quality data and correct the training error accumulation effectively.
By combining the CICL approach, a stable and competitive result is achieved, which demonstrate the approach is helpful for boosting the performance.
Finally, NTL is added to validate whether it can learn from the noise-free area in noisy labels. We observe that there is a great improvement in the case of large noise ratios.

\section{Conclusion}

This paper presented a novel cross-denoising framework, 
exploring the noisily annotated source domain images and unannotated target domain images  to improve the segmentation results of target images. 
In conjunction with a robust adversarial learning and a noise-tolerant loss, the domain shift and noisy labels problems can be solved simultaneously. Extensive experiments on OD and OC segmentation have demonstrated the advantages of our approach over the state-of-the-art alternatives. In addition to medical image, the method can also be valid for segmentation tasks where other types of images are not labeled accurately.

\section*{Acknowledegments}
This work was supported by the grants from Key Area Research and Development Program of Guangdong Province, China (No. 2018B010111001) and the Science and Technology Program of Shenzhen, China (No. ZDSYS201802021814180).
\bibliographystyle{named}
\bibliography{ijcai20}



\end{document}